%% file: main.tex
\documentclass[10pt,twocolumn,letterpaper]{article}

\usepackage[pagenumbers]{cvpr}


\definecolor{cvprblue}{rgb}{0.21,0.49,0.74}
\usepackage[pagebackref,breaklinks,colorlinks,allcolors=cvprblue]{hyperref}

\def\paperID{0000}
\def\confName{CVPR}
\def\confYear{2026}

\title{RePlan-Bot: Multi-Level Replanning for Embodied Instruction Following}

\author{
Xicheng Gong$^{1}$ \quad
Guozheng Sun$^{2}$ \quad
Peiran Xu$^{1}$ \quad
Yadong Mu$^{1,*}$\\[0.5ex]
$^{1}$Peking University \quad
$^{2}$Tsinghua University\\[0.5ex]
{\tt\small gongxicheng@stu.pku.edu.cn} \quad
{\tt\small sgz25@mails.tsinghua.edu.cn} \quad
{\tt\small xpr820@pku.edu.cn} \quad
{\tt\small myd@pku.edu.cn}\\[0.5ex]
{\small *Corresponding author}
}

\begin{document}

\maketitle

\input{sec/0_abstract}
\input{sec/1_intro}
\input{sec/2_formatting}
\input{sec/3_finalcopy}

{
    \small
    \bibliographystyle{ieeenat_fullname}
    \bibliography{main}
}


\end{document}

%% file: sec/0_abstract.tex
\begin{abstract}
Embodied instruction following (EIF) requires agents to understand and execute complex natural language commands within interactive 3D environments. Despite recent advances, existing methods often fail in long-horizon planning and handling irreversible state changes, resulting in low task success rates. To address these challenges, we introduce \textbf{RePlan-Bot}, a novel EIF agent that performs multi-level, continuous replanning throughout task execution. RePlan-Bot integrates a high-level LLM-based auditor for dynamic sub-goals adjustments guided by environmental feedback, a commonsense-guided search mechanism based on a multi-layered instance map for precise and structured object localization, and a lightweight ViT-based corrector to preemptively fix risky low-level actions. Evaluated on the ALFRED benchmark, RePlan-Bot achieves state-of-the-art performance in both seen and unseen environments, demonstrating superior adaptability and reliability.
\end{abstract}

%% file: sec/1_intro.tex
\section{Introduction}

Embodied instruction following (EIF) tasks require autonomous agents to understand and execute natural language commands within visually rich, interactive environments. Typical EIF scenarios, exemplified by benchmarks such as ALFRED~\cite{shridhar2020alfred}, involve sequences of navigation and manipulation actions guided by instructions like ``Clean the mug and place it on the countertop.'' These tasks pose significant challenges, as agents must handle long-horizon planning, cope with partial observability, and manage irreversible state changes. Consequently, even the most advanced EIF systems~\cite{song2023llm,min2021film,kim2023context,yang2024hindsight} often exhibit low task success rates, highlighting fundamental shortcomings of existing methods.

Previous EIF solutions~\cite{kim2021agent,pashevich2021episodic,suglia2021embodied,min2021film,murray2022following,inoue2022prompter,kim2023context,shin2024socratic} typically follow one of two strategies. End-to-end approaches directly map sensory inputs to actions, but frequently fail to generalize beyond training scenarios due to limited reasoning capability and insufficient interpretability; Modular methods, represented by systems such as FILM~\cite{min2021film} and CAPEAM~\cite{kim2023context}, decompose EIF tasks into high-level planning, mid-level object searching, and low-level action execution. While modularity improves interpretability, these approaches remain vulnerable because their modules typically lack adaptability: high-level planners produce static action templates disconnected from real-time visual feedback; mid-level search mechanisms use coarse semantic heatmaps inadequate for precise localization of small or partially occluded objects; and low-level controllers rely on fixed, primitive actions, susceptible to errors arising from inaccurate perception or faulty actuation.
\begin{figure}
    \centering
    \includegraphics[width=1\linewidth]{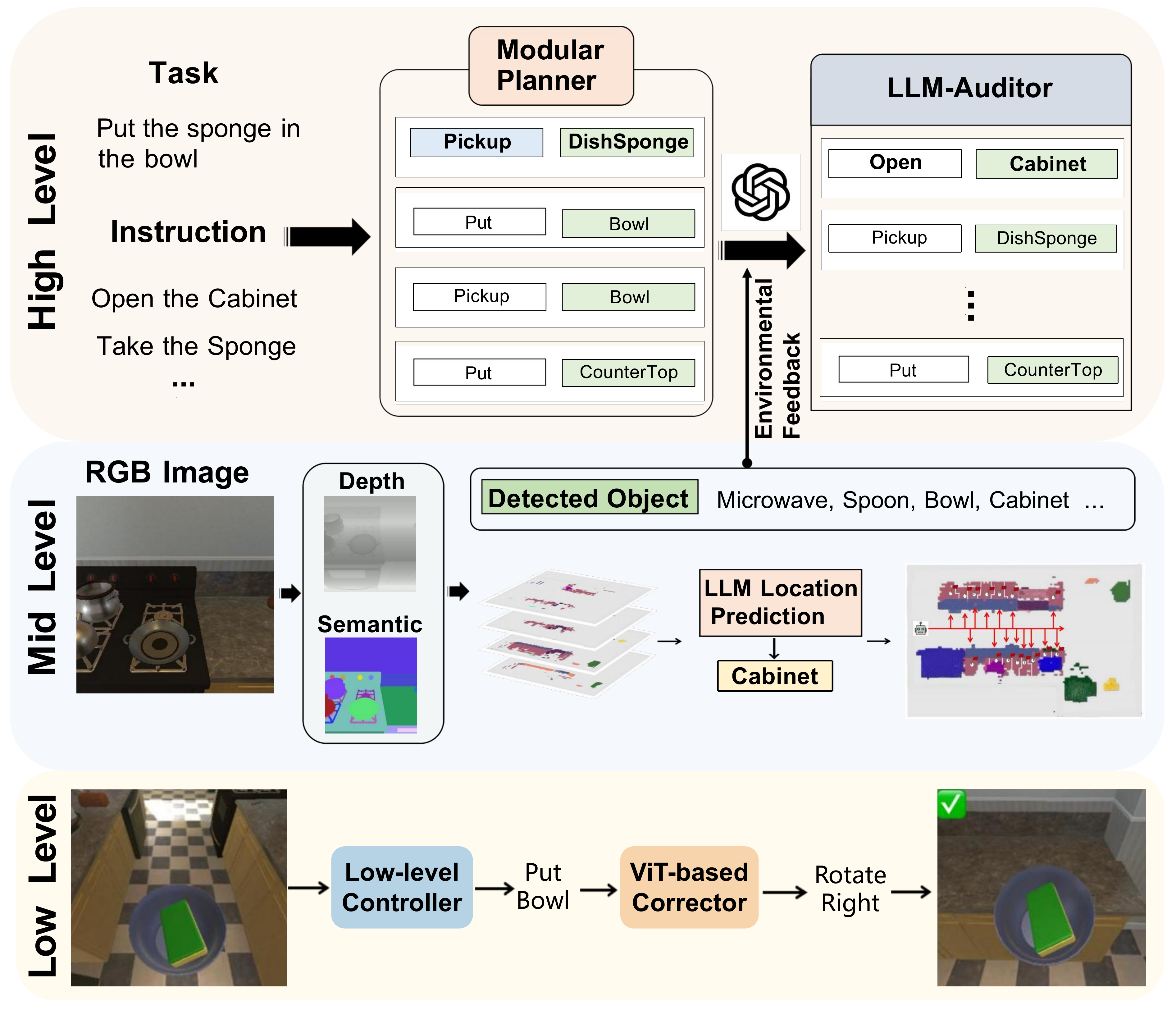}
    \caption{\textbf{Overview of the proposed RePlan-Bot}. It consists of three components: High Level Replanning, an LLM-based Auditor dynamically replans high-level goals; Mid Level Searching, a commonsense-driven module guides object search via multi-layered instance map; and Low Level Replanning, a ViT-based Corrector monitors low-level actions to prevent execution failures.}
    \label{fig:introduction}
\end{figure}
Addressing these critical limitations, we propose RePlan-Bot, a novel EIF agent that integrates multi-level and continuous replanning. As shown in Figure~\ref{fig:introduction}, at the high level, we introduce an environment-aware auditor powered by a large language model (LLM) that dynamically adjusts sub-goals by considering both textual instructions and current semantic context. For mid-level exploration, we present a commonsense-driven search mechanism leveraging a multi-layered instance map segmented by object instances and vertical spatial information, enabling targeted and structured exploration for unseen objects. At the low level, inspired by the intuition that humans continuously adjust their actions based on visual observations in the real world, we propose a lightweight vision transformer (ViT)-based ~\cite{dosovitskiy2020image} action corrector trained on a dataset constructed from execution trajectories. The corrector monitors each primitive action, evaluates its feasibility using egocentric images, and proactively substitutes risky actions with safer alternatives. This comprehensive framework significantly improves the robustness and overall performance of the EIF agent. Experimental results on the challenging ALFRED benchmark~\cite{shridhar2020alfred} demonstrate that RePlan-Bot substantially surpasses prior state-of-the-art methods, achieving marked improvements in both seen and unseen scenarios. Extensive ablations further confirm the individual and collective contributions of our proposed multi-level replanning approach.

Our main contributions can be summarized as follows:
\begin{itemize}

\item We design a \textbf{high-level environment-aware auditor} powered by LLM, which refines task plans on the fly by jointly leveraging natural language instructions and real-time environmental feedback.

\item We propose a \textbf{commonsense-guided search mechanism} based on a multi-layered instance map, enabling precise object localization and structured exploration, especially under partial occlusion or cluttered scenes.

\item We develop a \textbf{ViT-based low-level action corrector} trained on visual observations captured from execution trajectories, capable of predicting and correcting potential failures in primitive actions, thereby improving execution accuracy and safety.

\item We conduct extensive experiments on the ALFRED benchmark~\cite{shridhar2020alfred}, where RePlan-Bot achieves new state-of-the-art performance in both seen and unseen environments. Detailed ablations validate the effectiveness of each proposed component.
\end{itemize}

%% file: sec/2_formatting.tex
\section{Related Work}
\label{gen_inst}

\paragraph{Embodied Instruction Following}
Embodied Instruction Following (EIF) tasks require agents to translate natural-language commands into multi-step behaviors in 3D, partially observed environments. Early end-to-end approaches directly map egocentric observations and language to actions~\cite{kim2021agent,nguyen2021look,pashevich2021episodic,suglia2021embodied,song2022one,zhang2021hierarchical}. For example, EmBERT~\cite{suglia2021embodied} fuses panoramic RGB and step-wise text via a multimodal transformer to predict the next action. However, such models generalize poorly, relying heavily on  annotated data and often mis-grounding objects in cluttered scenes. 

To overcome these limitations, recent modular pipelines decouple high-level planning from low-level control~\cite{min2021film,kim2023context,murray2022following,ding2023embodied,inoue2022prompter,devlin2019bert,zhu2023ghost}, inferring abstract action plans and executing them via semantic or instance-level maps. FILM~\cite{min2021film} segments instructions into sub-goals and incrementally builds a semantic map, while CAPEAM~\cite{kim2023context} leverages a context-aware planner and memory for robust long-horizon manipulation.

\paragraph{LLMs for Embodied Instruction Following}
Large language models (LLMs) have emerged as powerful high-level planners in embodied AI, demonstrating strong generalization across tasks such as vision–language navigation~\cite{zhou2024navgpt,qiao2023march,chen2024mapgpt,lu2023multimodal,wu2023embodied,shah2023lm}, object-centric manipulation~\cite{yu2023l3mvn,yin2024sg,jeong2025objectcentricworldmodellanguageguided,wang2025instruction}, and long-horizon household planning~\cite{ahn2022can,raman2022planning,singh2023progprompt,lu2023thinkbot,chen2023robogpt,mu2023embodiedgptvisionlanguagepretrainingembodied,wu2024embodied}. Typically, LLMs generate sub-goals or action sketches from language instructions, and then low-level controllers handle execution. Despite strong zero-shot generalization via commonsense priors, LLMs lack explicit spatial grounding and struggle with fine-grained control due to large action spaces. To address this, hybrid architectures assign high-level reasoning to LLMs and delegate perception and control to specialized modules. For instance, ThinkBot~\cite{lu2023thinkbot} completes sparse instructions via self-reasoning and grounds plans using a multimodal object localizer.

On this basis, to address execution failures stemming from incomplete perception, ambiguous instructions, or dynamic environmental changes, recent methods integrate adaptive replanning modules into embodied agents~\cite{song2023llm,lu2023thinkbot,chen2023robogpt,shin2025socraticplannerselfqabasedzeroshot,kim2025multimodal,blukis2022persistent,wang2023describe,mei2024replanvlmreplanningrobotictasks,yoneda2024statlerstatemaintaininglanguagemodels,skreta2024replan}. This enhances robustness in long-horizon and partially observable tasks. For instance, FLARE~\cite{kim2025multimodal} substitutes undetected objects with semantically similar alternatives using language-encoder similarity (e.g., “TrashCan”→“GarbageCan”). These methods enable real-time and language-aware replanning, allowing agents to adapt fluidly in open-world scenarios.

\begin{figure*}
    \centering
    \includegraphics[width=1\linewidth]{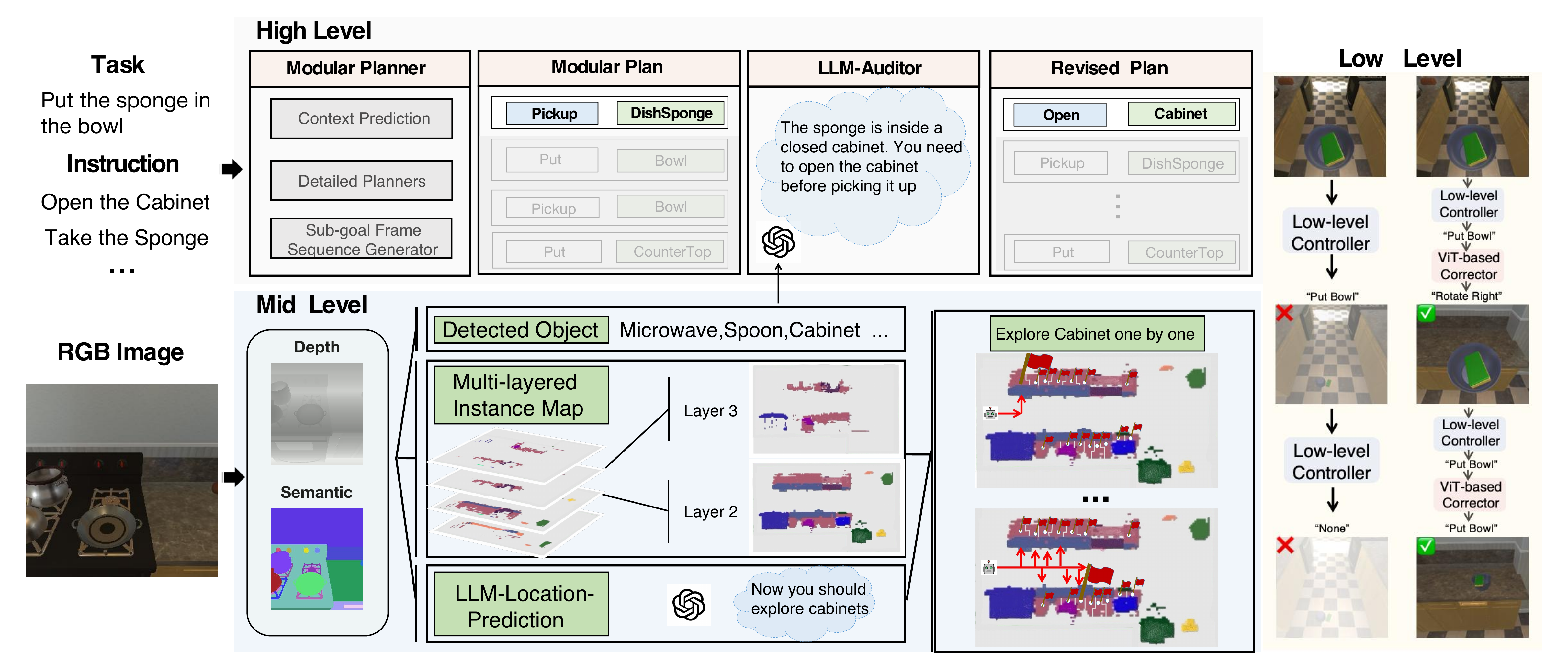}
    \caption{\textbf{The detailed pipeline of RePlan-Bot.} At the high level, upon receiving natural-language commands, the Modular Planner generates an initial plan. The LLM-Auditor then refines this plan to make it more rational. During task execution, the LLM-Auditor continuously optimizes the plan based on environmental feedback. At the mid level, the commonsense-guided search mechanism uses a multi-layered instance map and an LLM to predict object locations (“\raisebox{-0.2em}{\includegraphics[height=1em]{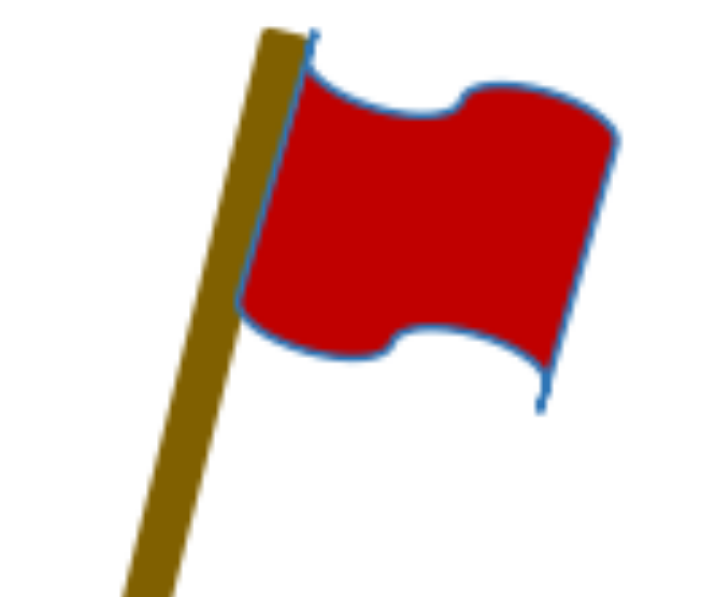}}”) for precise and efficient search. At the low level, the ViT-based corrector adjusts the originally planned low-level actions based on the current RGB input to ensure correct execution.}
    \label{fig:2}
\end{figure*}

%% file: sec/3_finalcopy.tex
\section{Approach}
\label{headings}

The typical EIF pipeline~\cite{min2021film,kim2023context}  struggles when language is sparse or ambiguous, which often results in logically incomplete plans. In addition, single-layer semantic map routinely misses small or occluded items. Moreover, rigid controller that ignores ongoing perceptual feedback cannot predict or recover from execution failures, leading to aimless exploration, collisions, and cascading errors.

To overcome these limitations, we introduce a continuous multi-level replanning approach, as shown in Figure~\ref{fig:2}. At the high level, an LLM-based auditor refines under-specified sub-goal sequences by integrating instruction and environmental feedback, enabling real-time adjustment to dynamic scenes. At the mid level, a commonsense-guided search mechanism leverages a multi-layered instance map, which enables thorough and efficient exploration, particularly in environments containing multiple instances of the same object category. At the low level, a ViT-based corrector continuously monitors egocentric observations to anticipate failures and proactively substitute risky actions with safer alternatives to enhance overall task reliability. This hierarchical replanning mechanism enables the agent to revise flawed high-level plans, navigate purposefully under partial observability, and recover robustly from execution failures. Below is the detailed explanation of each component.

\subsection{High-Level Replanning: Environment-Aware Sub-goals Auditing}
Large language models excel at inferring latent structure from language and chaining symbolic steps, both of which naturally align with the high-level planning demands of embodied instruction following~\cite{song2023llm,inoue2022prompter}. We exploit these strengths by inserting an LLM auditor after the Modular High-Level Planner. The auditor receives a structured prompt containing three streams of information: System Message, including the role explanation, task definition, and response format; Agent Message, which indicates the task instruction and the given sparse step-by-step instruction sentences; Environmental Feedback, which includes the observed objects and the current step index in sub-goals list. It must emit a revised sub-goals list that conforms to the same schema yet is fully grounded in the observed scene. We detail the language model setup and provide the full prompt structure in Appendix 1.2.

Specifically, high-level replanning is triggered under these two conditions: (1) at the beginning of each task, an initial audit is introduced to ensure that the sub-goals list is consistent with the scene configuration and task instruction; (2) when the agent takes an excessive number of steps to complete a sub-goal, the auditor re-evaluates the sub-goal in context to stay aligned with the evolving environment.
This hybrid trigger ensures that replanning is both proactive and anticipatory, making it capable of adapting in real time. The details are described as follows.

\subsubsection{Pre-Execution Sub-Goals Refinement}

In many modular approaches such as CAPEAM~\cite{kim2023context}, high-level sub-goals are generated by matching objects with predefined action templates, without reasoning about their spatial state. This may produce infeasible plans: for instance, the agent may attempt (PickUp, DishSponge) even when the object is inside a closed cabinet, without opening it first.

To address this limitation, we introduce an LLM-based auditor that refines the sub-goals prior to the execution of each task. At the beginning of each task, the auditor performs an initial review of the sub-goals list generated by the Modular High-Level Planner, supplementing missing actions and correcting object reference errors arising from semantic ambiguity. For instance, in the above case, the auditor augments the plan with (Open, Cabinet) before attempting the pickup. By explicitly reasoning over spatial visibility and containment, the auditor identifies and corrects missing preconditions that would otherwise cause execution to fail.

\subsubsection{Vision-Grounded Object Disambiguation}
Existing instruction-following systems often select object references purely based on textual similarity, without grounding them in the visual context. This can result in selecting semantically relevant but visually absent targets, leading to invalid or unnecessary search behaviors, especially in cluttered or visually ambiguous environments.

To address this issue, we leverage the LLM-based auditor to refine sub-goals based on the detected-object set and the original sub-goals list. For instance, when the agent takes too many steps to execute the action (PickUp, Cup), the auditor may revise this sub-goal to (PickUp, Mug), guided by the fact that Mug is present in the detected object set. This adjustment ensures successful task completion. By grounding symbolic references in actual visual observations, the auditor substitutes unsupported predictions with visually available alternatives. This alignment between symbolic planning and egocentric perception enhances the robustness of the overall plan and reduces failures caused by semantic misalignment.

\subsection{Mid-Level Searching: Commonsense-Guided Search Mechanism}
\begin{figure}
    \centering
    \includegraphics[width=1\linewidth]{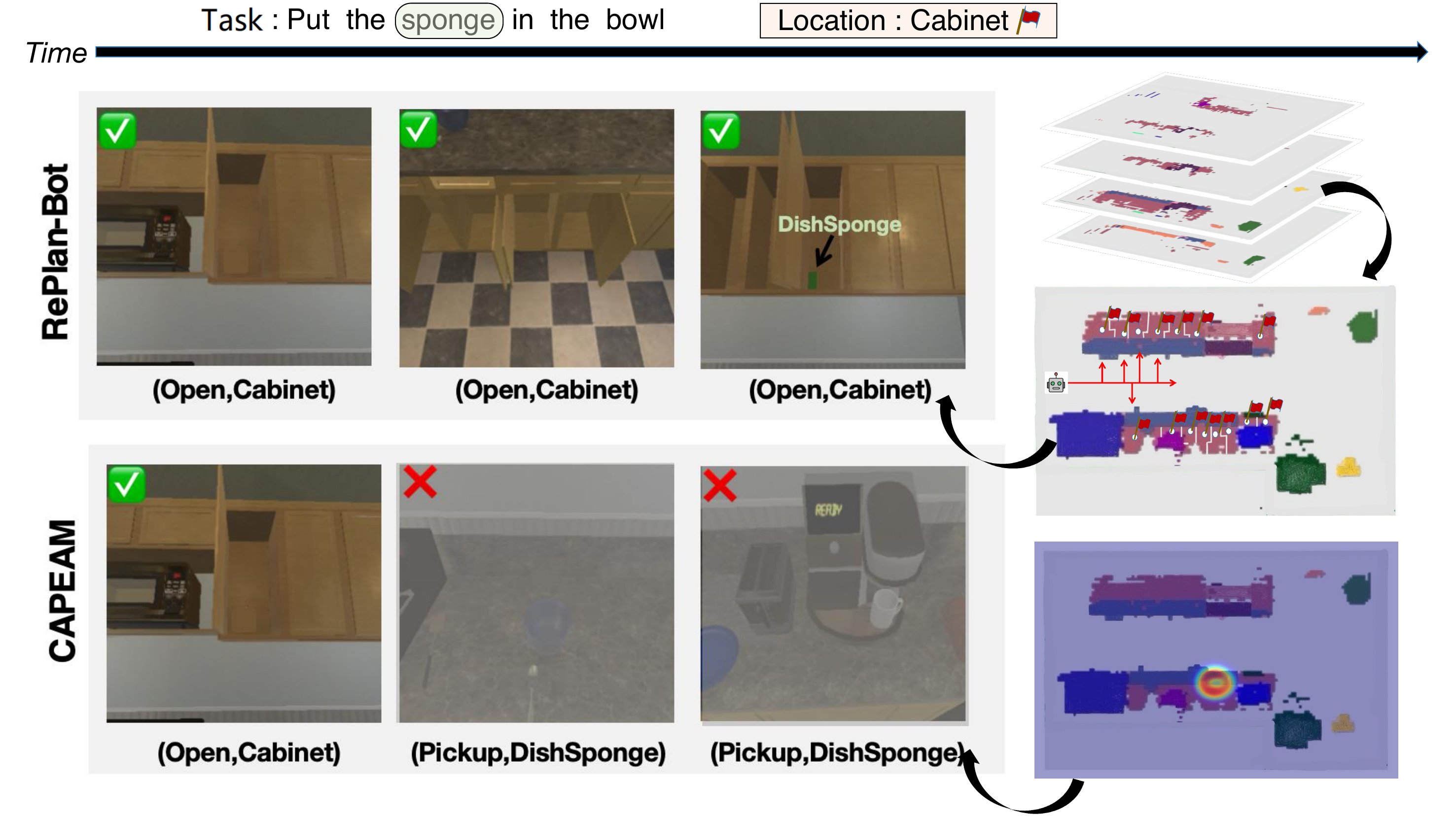}
    \caption{\textbf{Comparison between RePlan-Bot and conventional EIF methods (CAPEAM~\cite{kim2023context}).} RePlan-Bot predicts the sponge is in a cabinet (“\includegraphics[height=1em]{flag.png}”) and successfully finds it after exploring multiple cabinets. In contrast, CAPEAM checks only one cabinet and moves on without verifying the sponge's location, resulting in task failure.}
    \label{fig:mid_level}
\end{figure}

Given a global semantic map $M_t \in \{0, 1\}^{C \times H \times W }$ and a subgoal-defined target object $o^\star$, the agent must determine where to begin its search when $o^\star$ is unobserved. Prior approaches such as FILM~\cite{min2021film} and CAPEAM~\cite{kim2023context} compress $M_t$ into a fixed-resolution heatmap $H_t \in \mathbb{R}^{8\times8}$ using a trained U-Net~\cite{ronneberger2015u}, which guides the agent toward a coarse likelihood peak. However, such predictions are spatially imprecise and often unreliable for small or novel items.

To facilitate a more precise and object-aware exploration search strategy, we construct a \textbf{multi-layered instance map} $M_t \in \{0, 1, \ldots, N\}^{C \times H \times W \times Z}$, where $C$ denotes the number of semantic categories such as \texttt{Cabinet} and \texttt{Fridge}, $H$ and $W$ represent the height and width of the map grid respectively, and $Z$ indicates discrete height slices obtained by partitioning the 3D voxelized observations vertically; meanwhile, $\{0, 1, \ldots, N\}$ correspond to the instance IDs assigned to each pixel within the same semantic category. At each timestep, the RGB image is processed into a depth map and instance segmentation mask using Mask R-CNN~\cite{he2017mask}, from which we recover a 3D point cloud. Each 3D point is associated with a predicted semantic category and discretized into a voxel grid along the vertical axis. 
This mapping strategy enables object localization across the scene layout while supporting instance-level differentiation along the vertical axis, such as distinguishing upper and lower cabinet compartments.

To update the multi-layered semantic map, we project the 3D voxelized observations into the appropriate height slice based on their vertical coordinates. Each pixel is labeled with an instance \textit{ID}, and contributes to the corresponding $(c, z)$ slice of the map. In each slice, we compute : 

\begin{footnotesize}
 \begin{align}
ID(I_t) = 
\begin{cases}
\arg\max\limits_{i \in \mathcal{I}_{t-1}} \mathrm{IoU}(I_t, i), & \text{if } \max \mathrm{IoU}(I_t, i) > \theta \\[4pt]
\mathrm{NewID}, & \text{otherwise}
\end{cases}
\end{align} 
\end{footnotesize}
where $i$ indexes a candidate instance in the maintained instance set $\mathcal{I}_{t-1}$, $\mathrm{IoU}(I_t, i)$ denotes the intersection-over-union between the current mask $I_t$ and instance $i$, and $\theta$ is a predefined threshold that determines whether the current mask $I_t$ sufficiently overlaps with any existing instance.
This process enables instance-level tracking conditioned on both semantic and height-based spatial constraints.

Building upon this spatially grounded map, we now describe how the agent performs target-directed search. When the target object is small or visually occluded, direct localization from egocentric perception is often unreliable. Humans naturally resolve this challenge by using semantic priors, inferring object locations based on their typical co-occurrence with larger, stable support objects. Inspired by this, we introduce a LLM that predicts a high-level host category $h^\star$ given the task description, step-by-step instruction, and the set of currently detected objects. The prediction acts as a semantic anchor to narrow the search space, while visited locations are dynamically masked to avoid redundancy. This targeted search strategy is triggered only when the goal object has not yet been observed in the environment. The language model configuration, along with detailed listings of small object classes and host category definitions, are presented in Appendix 1.3.

Once $h^\star$ is predicted, the agent queries the multi-layered instance map to enumerate all instances of the host category. Instead of relying on coarse likelihood heatmaps, the agent conducts a structured, instance-wise search by explicitly visiting each candidate in sequence. As shown in Figure \ref{fig:mid_level}, based on the LLM’s prediction that the sponge is inside a \texttt{Cabinet}, the agent explores multiple cabinet instances one by one and successfully locates the hidden sponge, which remains fully occluded throughout. By aligning language-driven semantic inference with spatial instance reasoning, our approach yields robust performance in cluttered and partially observable environments. 

\subsection{Low‐Level Replanning: Vision‐Driven Action Correction}

In most modular EIF systems, once the FMM planner~\cite{sethian1996fast} issues a short-term goal, the low-level controller adjusts its heading using basic “MoveAhead” and “Rotate” actions based on the semantic map. Upon reaching the goal region, it checks for a segmented mask of the target in the current RGB frame and executes a predefined action, such as PickUp, without judging whether the action is feasible. As a result, while executing actions, this template‐driven approach struggles to handle real‐world imperfections: the collision map may misrepresent free space due to coarse resolution; segmentation may miss the object under occlusion; and actuation errors such as grasp misalignment can send the agent off course. As a result, fixed primitives often attempt infeasible moves, such as driving into undetected obstacles, over-rotating, or grasping at empty air, which leads to repeated failures and wasted time.
\begin{figure}
    \centering
    \includegraphics[width=1\linewidth]{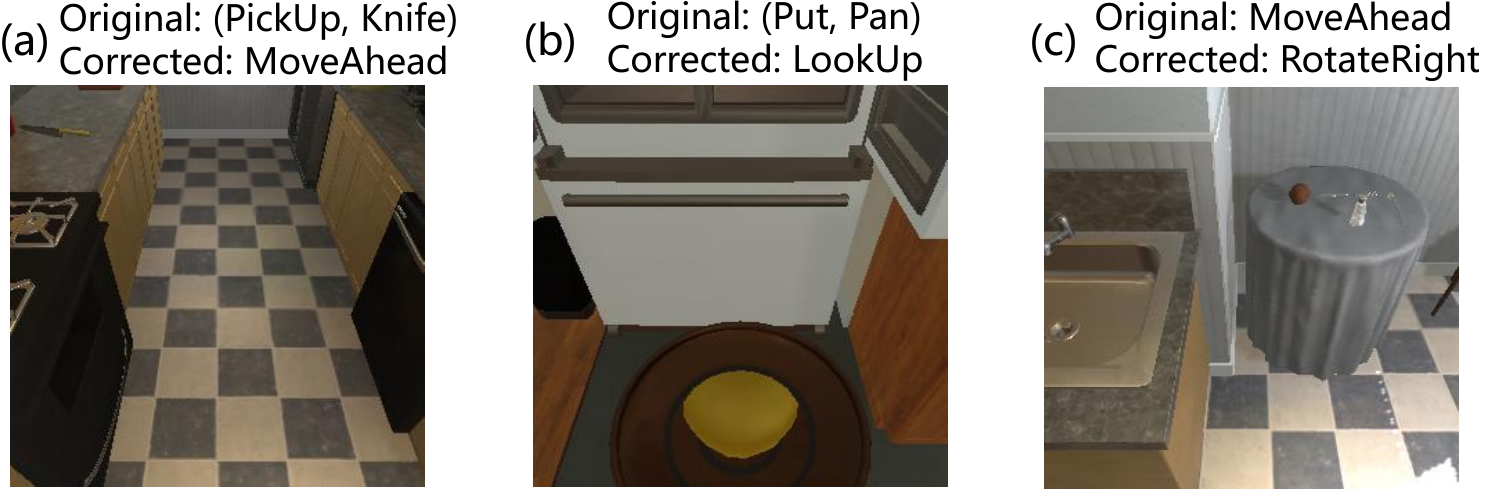}
    \caption{\textbf{Example visualization of the RGB image and the corresponding low-level action.} (a) The agent is too far to \texttt{PickUp Knife}, so the action is corrected to \texttt{MoveAhead}. (b) The viewpoint is too low to \texttt{Put Pan in Fridge}, so the action is corrected to \texttt{LookUp}. (c) The agent is blocked when trying to \texttt{MoveAhead}, so the action is corrected to \texttt{RotateRight}.
    }
    \label{fig:111}
    
\end{figure}

In the real world, humans continuously adjust their actions based on current visual information. Inspired by this, we interleave a lightweight ViT‐based corrector into each primitive step. Specifically, the ViT module takes the current egocentric RGB image and the originally planned low-level action as input, evaluates its feasibility, and outputs a potentially better alternative. Formally, we denote the RGB observation as $O_t$ at time $t$, and $a_t^{\text{plan}}$ the action proposed by the low-level controller. The ViT model computes:

\begin{equation}
\hat{a}_t = \mathcal{F}_{\text{ViT}}(I_t, a_t^{\text{plan}}),
\end{equation}
where $\mathcal{F}_{\text{ViT}}$ is a transformer-based policy head that outputs a corrected action $\hat{a}_t\in\mathcal{A}$, with $\mathcal{A}$ denoting the discrete low-level action space (e.g., \texttt{MoveAhead}, \texttt{RotateLeft}, \texttt{RotateRight}, \texttt{LookDown}, and \texttt{LookUp}). When $\hat{a}_t = a_t^{\text{plan}}$, the agent proceeds with the original plan; otherwise, the system treats $\hat{a}_t$ as a more suitable alternative and executes it instead.

To train the ViT module, each training sample consists of an RGB image $O_t$ and the planned action $a_t^{\text{plan}}$, where the action is embedded as a discrete token and concatenated with the image tokens before being passed to the transformer encoder. A softmax classification head predicts the corrected action over the space $\mathcal{A}$, supervised by the relabeled ground truth. This setup enables the ViT to function both as an action validator and a corrector, learning to recognize execution failures and predict better alternatives in visually ambiguous or error-prone situations. The training dataset and details are presented in Appendix 1.4.

\section{Experiments}
\label{others}
\subsection{Experimental Setup}
\paragraph{Dataset.}
We evaluate our method on the ALFRED benchmark~\cite{shridhar2020alfred}, which contains 25,743 instruction-demonstration pairs for interactive 3D household tasks in AI2-THOR~\cite{kolve2017ai2}. The dataset spans seven task types involving long-horizon, multi-step navigation and manipulation under partial observability. It is split into training, validation, and test sets, with validation and test environments further divided into seen and unseen folds to assess generalization.

\paragraph{Evaluation Metrics.}
We report official ALFRED metrics: Success Rate (SR, percentage of episodes with all sub-goals completed), Goal-Condition Success Rate (GC, proportion of individual goal conditions achieved), and their path length-weighted variants (PLWSR, PLWGC), which penalize inefficient trajectories. These metrics jointly evaluate both the effectiveness and efficiency of instruction following. Additional details are provided in Appendix 1.1.
\subsection{Comparison with the State of the Art}

\begin{table*}[t]
\footnotesize
\centering
\caption{Comparison with state-of-the-art methods on the ALFRED benchmark (Test Seen/Unseen).}
\label{tab:sota}

\begin{tabular}{lcccc}
\toprule
\textbf{Method} & \multicolumn{2}{c}{\textbf{Test Seen}} & \multicolumn{2}{c}{\textbf{Test Unseen}} \\
\cmidrule(lr){2-3} \cmidrule(lr){4-5}
 & GC(PLWGC) & SR(PLWSR) & GC(PLWGC) &  SR(PLWSR) \\
\midrule
\multicolumn{5}{c}{\textit{Task Description Only}} \\
HLSM    & 35.79(11.53) & 25.11(6.69)  & 27.24(8.45)   & 16.29(4.34) \\
FILM             & 36.15(14.17) & 25.77(10.39) & 34.75(13.13) & 24.46(9.67) \\
LGS-RPA          & 41.71(24.49) & 33.01(16.65) & 38.55(20.01) & 27.80(12.92) \\
Prompter         & 56.98(\textbf{28.42}) & 47.95(23.29) & 56.57(25.80) & 41.53(18.84) \\
CAPEAM           & 53.14(24.32) & 43.64(19.76) & 54.17(23.68) & 40.42(19.34) \\
\textbf{RePlan-Bot} &\textbf{57.81}(25.23)& \textbf{49.23}(\textbf{23.37}) & \textbf{57.02}(\textbf{26.31}) & \textbf{44.79}(\textbf{21.89}) \\
\midrule
\multicolumn{5}{c}{\textit{Task Description + Step-by-step Instructions}} \\
ABP   & 51.13(4.92) & 44.55(3.88)  & 24.76(2.22)   & 15.43(1.08) \\
FILM             & 39.55(15.59) & 28.83(11.27) & 38.52(15.13) & 27.80(11.32) \\
LGS-RPA          & 48.66(28.97) & 40.05(21.28) & 45.24(22.76) & 35.41(\textbf{22.76}) \\
Prompter         & 60.22(\textbf{30.21}) & 51.17(\textbf{25.12}) & 56.57(25.80) & 45.32(20.79) \\
CAPEAM           & 57.32(26.31) & 47.61(22.82) & 56.52(24.26) & 43.17(20.13) \\
\textbf{RePlan-Bot} & \textbf{61.21}(26.69) & \textbf{52.05}(24.60) & \textbf{60.29}(\textbf{26.31}) & \textbf{47.61}(21.89) \\
\bottomrule
\end{tabular}
\end{table*}

We compare RePlan-Bot with several recent state-of-the-art models on the ALFRED benchmark, including FILM~\cite{min2021film}, LGS-RPA~\cite{murray2022following}, Prompter~\cite{inoue2022prompter}, CAPEAM~\cite{kim2023context}, and so on. Since our approach is based on CAPEAM, we use the reproduced results for CAPEAM in our evaluation, while the results for other baselines are taken from the reported values. Notably, the reproduced results of CAPEAM differ slightly from those originally reported~\cite{kimpre}. All methods are evaluated with task description only and with both task description and step-by-step instructions as input, using PLWGC, GC, PLWSR, and SR on both Test Seen and Test Unseen splits.
\paragraph{Task Description Only.}
In the setting where only the task description is provided, RePlan-Bot achieves 44.79\% SR and 57.02\% GC on the Test Unseen split, outperforming all baselines by a clear margin (e.g., CAPEAM: 40.42\% SR, 54.17\% GC). On the Test Seen split, our method also leads with 49.23\% SR and 57.81\% GC. This substantial improvement demonstrates strong generalization to novel environments, even with limited high-level instructions. Such robustness makes RePlan-Bot practical for real-world scenarios where step-by-step guidance may be unavailable.

\paragraph{Both Task Description and Step-by-Step Instructions.}
When both the task description and step-by-step instructions are provided, RePlan-Bot further improves across both SR and GC, reaching 47.61\% SR and 60.29\% GC on the Test Unseen split, and 52.05\% SR and 61.21\% GC on the Test Seen split. These results highlight the model's ability to leverage detailed instructions for more accurate and complete task execution, which indicates that our approach is both robust to instruction sparsity and highly efficient in utilizing additional information.


\begin{figure}
    \centering

    \includegraphics[width=1\linewidth]{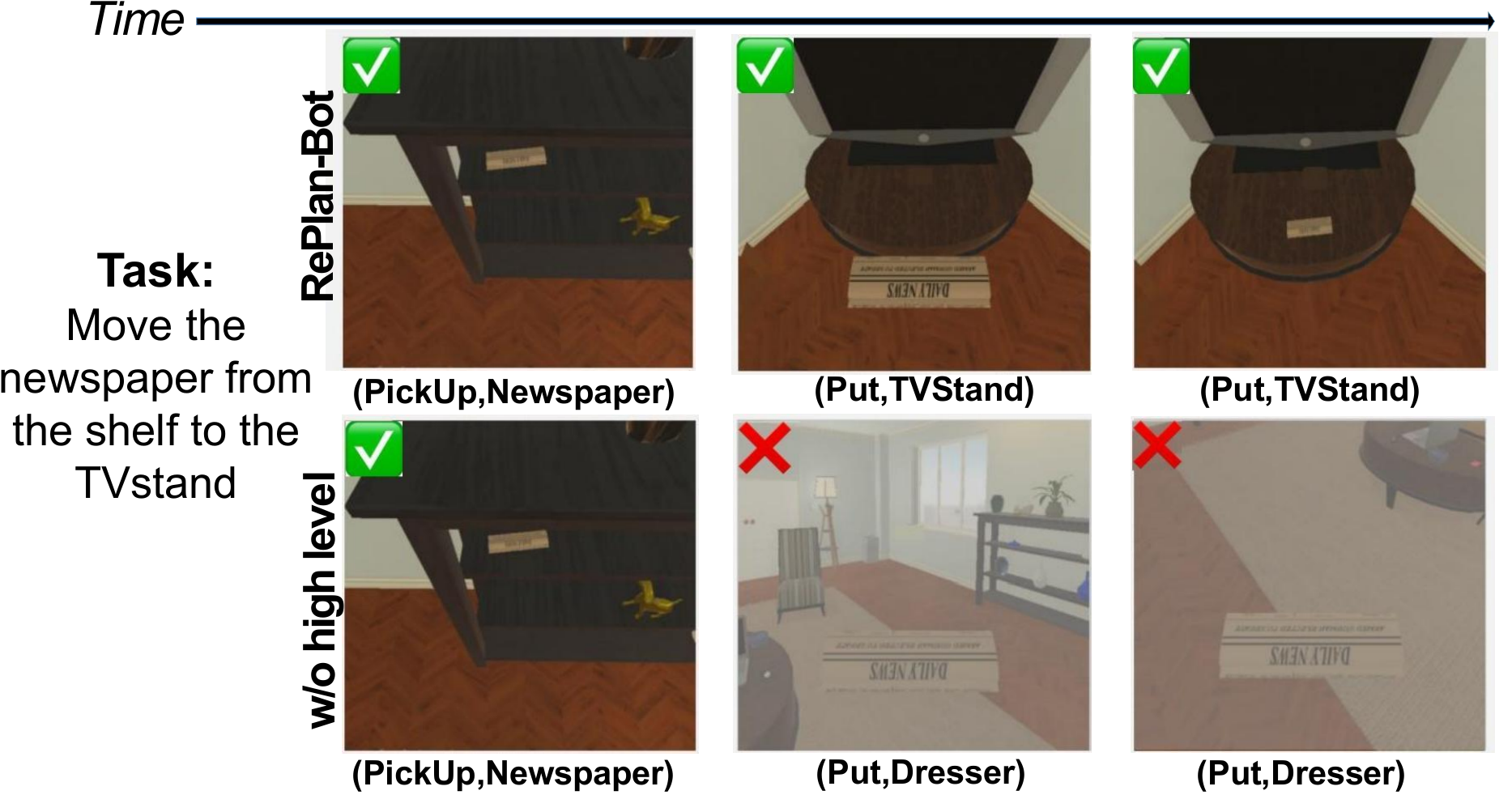}
    \caption{\textbf{Comparison between RePlan-Bot with and without high-level replanning.} The top row shows correct actions guided by high-level replanning, while the bottom row shows failed actions without it.}
    \label{fig:high_level_appendix}
\end{figure}

\subsection{Ablation Study}
To evaluate the contributions of each proposed component in our RePlan-Bot, we conduct a quantitative ablation study, with the results summarized in Table~\ref{tab:ablation_study}, where both the task description and step-by-step instructions are provided. 

\paragraph{Without High-Level Replanning.}
Removing the high-level replanner consistently degrades performance: SR and GC are reduced by 3.19\% and 2.40\%, respectively, on the Test-Seen split, and by 2.87\% and 2.42\% on Test-Unseen. Without this layer, the agent tends to generate incomplete sub-goals or mis-referenced objects. High-level replanning promptly rectifies such errors, ensuring that actions remain aligned with task instructions and real-time environmental feedback throughout the episode. This global, adaptive oversight is essential for robust and efficient long-horizon execution, distinguishing our approach from conventional modular planners.
\begin{figure}
    \centering
    \includegraphics[width=1\linewidth]{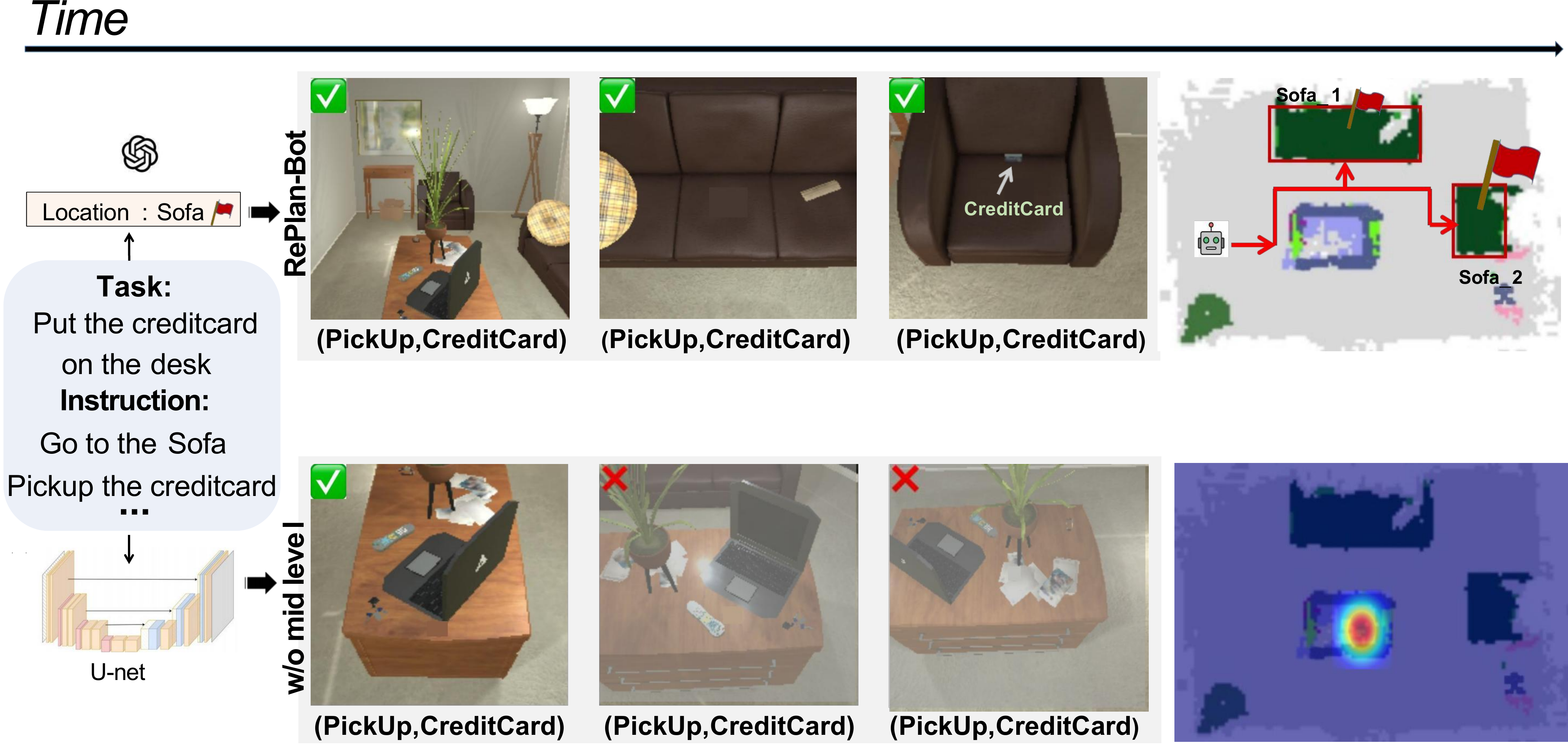}
    \caption{\textbf{Comparison between RePlan-Bot with and without mid-level searching.} The top row shows correct actions guided by commonsense-guided search mechanism, while the bottom row shows failed actions based on Unet.}
    \label{fig:appendix_2}
\end{figure}

\paragraph{Without Mid-Level Searching.}

Table~\ref{tab:ablation_study} shows that removing the mid-level searching module leads to a noticeable performance drop on the Test Unseen set, where the SR decreases by 1.70\% and GC drops by 0.31\%. Although the decline on the Test Seen set is less pronounced (SR drops by 0.91\% and GC by 0.28\%), the sharper degradation on unseen scenarios highlights the importance of mid-level reasoning when generalizing to unfamiliar environments. Meanwhile, in real-world scenarios where partial observability, occlusions, and cluttered scenes are much more prevalent, the mid-level search mechanism may be essential for reliably locating and distinguishing objects.

\paragraph{Without Low-Level Replanning.}
Table~\ref{tab:ablation_study} indicates that removing the low-level action corrector leads to a clear drop in all metrics. Specifically, the SR decreases by 2.08\% and GC drops by 2.19\% on the Test Seen set; on the Test Unseen set, SR decreases by 2.41\% and GC drops by 1.94\%. This demonstrates that although predefined actions can execute simple plans, they are prone to failure in the absence of visual feedback. The vision-based corrector allows the agent to dynamically adjust actions in response to current visual feedback, preventing repeated failures such as unsuccessful pickup or unnecessary collisions. As a result, the agent achieves more efficient and robust task execution, reflected by notable improvements over both the baseline and the ablated variant in all evaluation settings.

\subsection{Qualitative Analysis}

\paragraph{High-Level Replanning}

Figure~\ref{fig:high_level_appendix} qualitatively demonstrates the effectiveness of our high-level replanning module on representative instruction-following tasks. In the “chill a tomato and put it inside the microwave” scenario, the baseline agent without high-level replanning keeps pushing the tomato against a closed microwave and never opens the door. In contrast, our method employs an LLM auditor that reviews the initial plan, adds essential missing steps including opening the microwave and selecting the correct object, and aligns the revised sequence with the observed scene before execution. These qualitative cases highlight that high-level replanning is essential for robust execution in complex environments, especially when instructions contain sequential dependencies or linguistic ambiguity.

\begin{figure}
    \centering
    \includegraphics[width=1\linewidth]{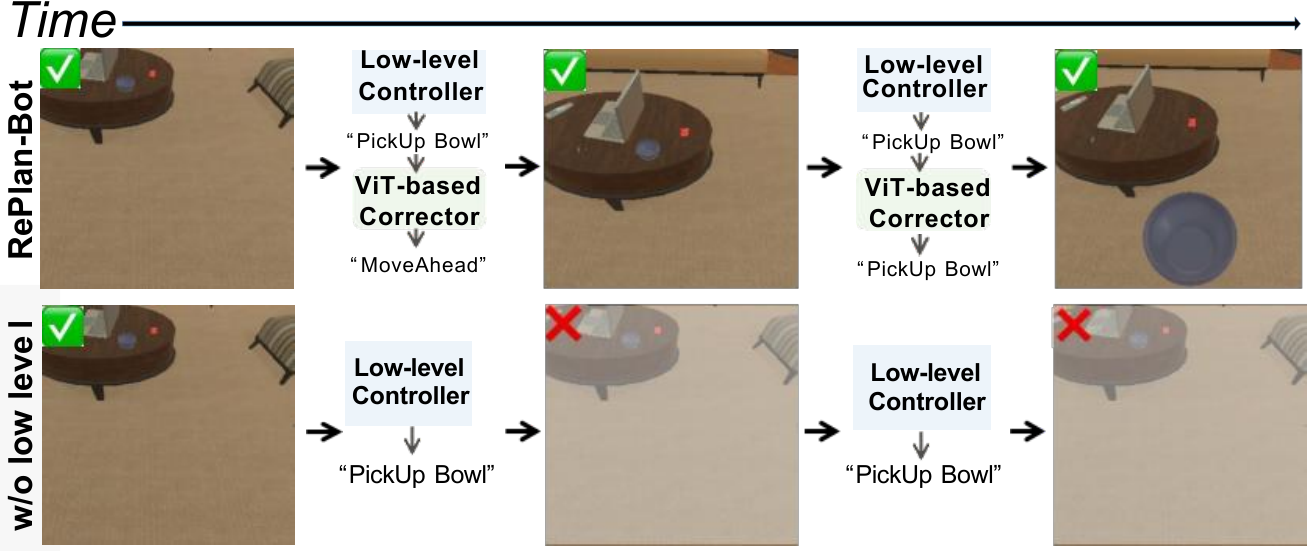}
    \caption{\textbf{Comparison between RePlan-Bot with and without low-level replanning.} With the low-level action corrector, the agent successfully picks up the bowl on the table. Without the low-level action corrector, the agent fails to do so due to being positioned too far from the table.}
    \label{fig:appendix3}
\end{figure}
\begin{table}[t]
\centering

\setlength{\tabcolsep}{2pt} 
\resizebox{\linewidth}{!}{%
\begin{tabular}{lcccc}
\toprule
\textbf{Method} & \multicolumn{2}{c}{\textbf{Test Seen}} & \multicolumn{2}{c}{\textbf{Test Unseen}} \\
\cmidrule(lr){2-3} \cmidrule(lr){4-5}
 & GC(PLWGC) & SR(PLWSR) & GC(PLWGC) &  SR(PLWSR) \\

\midrule
\textbf{RePlan-Bot}   & 61.21(26.69)    & 52.05(24.60)   & 60.29(26.31)  & 47.61(21.89) \\
\midrule
w/o High-Level  & 58.81(25.10)   & 48.86(22.63)    & 57.87(24.33)   & 44.74(20.24) \\
w/o Mid-Level  & 60.93(26.54)    & 51.14(23.11)    & 59.98(25.64)   & 45.91(20.95) \\
w/o Low-Level & 59.02(24.79)    & 49.97(21.76)    & 58.35(23.93)   & 45.20(19.22) \\
\bottomrule
\end{tabular}
}
\caption{Ablation study for each proposed component.}
\label{tab:ablation_study}
\end{table}
\paragraph{Mid-Level Searching}
Figure \ref{fig:appendix_2} highlights the value of our commonsense-guided mid-level search for object retrieval. Charged with placing a credit card on the desk, the agent first receives the language model’s guess that a sofa may host the card. RePlan-Bot then consults its instance-level semantic map and inspects each sofa in sequence, quickly uncovering the hidden card and completing the task (top row). Without the mid-level module, the ablated agent relies on a U-Net likelihood heat map whose coarse spatial cues misdirect the search and leave the card unfound. The example shows that instance-wise exploration delivers greater precision and robustness than heat-map methods, particularly for small or occluded objects.

\paragraph{Low-Level Replanning}
Figure~\ref{fig:appendix3} presents an example where our ViT-based low-level action corrector significantly improves execution success rates compared to the baseline without the corrector. 
For the ``Put Bowl'' action, the corrector similarly recommends corrective moves to ensure proper placement, while the baseline often fails due to poor positioning or actuation errors.
These results highlight that our vision-driven low-level replanning provides crucial robustness against spatial misalignment, sensor noise, and control errors, enabling more reliable object manipulation than naive template-based controllers.

\subsection{Application in Generalizable Robotic Manipulation}

\begin{figure}
    \centering
    \includegraphics[width=1\linewidth]{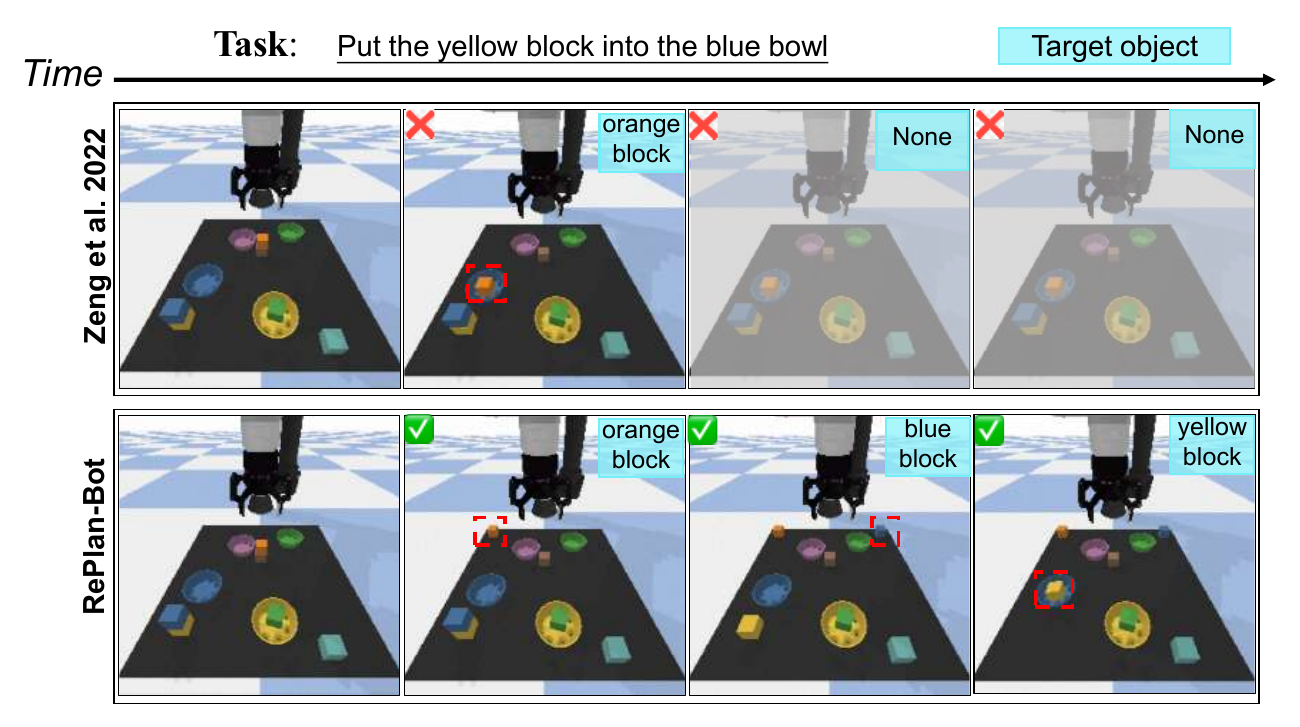}
    \caption{\textbf{An example of multi-step planning for a complex manipulation task}. The baseline model ~\cite{zeng2022socraticmodels} fails due to a static plan that cannot handle the occluded target object. In contrast, RePlan-Bot formulates a dynamic plan to first clear the occluding blocks and successfully completes the task.}
    \label{fig:mani}
\end{figure}

To demonstrate the generalization of RePlan-Bot beyond the ALFRED benchmark,  we use a simulated tabletop environment and illustrate a comparison between RePlan-Bot and the baseline model in Figure~\ref{fig:mani}. We choose ~\cite{zeng2022socraticmodels} as the baseline model for its effectiveness in few-shot robot planning. A common methodological ground is shared between the two systems, as both utilize GPT-4o as a high-level LLM for sub-goal generation and rely on an identical privileged low-level policy for execution.
Additional details are provided in Appendix 1.5.

We observe that RePlan-Bot successfully completes the complex pick-and-place task, demonstrating its ability to solve occlusion problems through high-level replanning and mid-level searching. In contrast, the baseline model fails as its static plan is incapable of generating the necessary sub-goals to first move the occluding orange and blue blocks to access the target yellow block.
\section{Conclusion}
In this paper, we propose \textbf{RePlan-Bot}, a novel EIF agent that integrates multi-level, continuous replanning to address long-horizon planning and complex environment challenges. The high-level LLM-based auditor dynamically adjusts sub-goals using environmental feedback, while the mid-level commonsense-guided search mechanism enables precise object localization with multi-layered instance maps. At the low level, the ViT-based corrector preemptively fixes risky actions using visual observations. Evaluated on the ALFRED benchmark, RePlan-Bot delivers exceptional performance across both seen and unseen environments, showcasing its strong adaptability and robustness. 

\paragraph{Limitations and Future Work}
In real-world applications, agents often encounter unexpected events or adversarial conditions, which are not fully reflected in current simulation benchmarks. As a result, bridging the sim-to-real gap remains a significant challenge, so our future work may explore integrating reinforcement learning, domain adaptation, or more realistic simulators to enhance the agent’s robustness in real-world scenarios.